%% file: acl_latex.tex
\documentclass[11pt]{article}

\usepackage[preprint]{acl}

\usepackage{times}
\usepackage{latexsym}

\usepackage[T1]{fontenc}
\usepackage[utf8]{inputenc}

\usepackage{microtype}

\usepackage{url}
\usepackage{dblfloatfix}
\usepackage{amsmath}
\usepackage{threeparttable}
\usepackage{booktabs}
\usepackage{makecell}
\usepackage{multirow}
\usepackage{arydshln}
\usepackage{adjustbox}
\usepackage{amssymb}
\usepackage{caption}
\usepackage{tabularx}
\usepackage{siunitx}
\newcommand{\sys} {RC-RAG} % {NSymBrid}
\newcommand{\gpt}{GPT-4o mini}
\newcommand{\llamaSevenB}{Llama-2-7b-chat-hf}
\newcommand{\llamaThirteenB}{Llama-2-13b-chat-hf}
\newcommand{\llamaEightB}{Meta-Llama-3.1-8B-Instruct}
\newcommand{\mistral}{Mistral-7B-Instruct-v0.3}
\newcommand{\mixtral}{Mixtral-8x7B-Instruct-v0.1}

\usepackage{inconsolata}

\usepackage{graphicx}

\title{Bridging the Long-Tail Gap: Robust Retrieval-Augmented Relation Completion via Multi-Stage Paraphrase Infusion}

\author{
Fahmida Alam, Mihai Surdeanu, Ellen Riloff \\
Department of Computer Science\\
University of Arizona, USA \\
\texttt{\{fahmidaalam, msurdeanu, riloff\}@arizona.edu}
}

\begin{document}
\maketitle

\begin{abstract}
\input{tex/abstract}

\end{abstract}

\section{Introduction}
\input{tex/introduction}

\input{tex/RAGImage}
\section{Related Work}
\label{sec:related_work}
\input{tex/related_work}

\section{Proposed Method}
\input{tex/proposed_method}

\section{Experimental Setup}
\label{sec:experimental_setup}
\input{tex/experiments}

\section{Results and Analysis}
\label{sec:result_analysis}
\input{tex/result_and_analysis}

\section{Conclusion}
\input{tex/conclusion}

\section*{Limitations}
\input{tex/limitation}

%\section*{Acknowledgments}

\bibliography{custom}

\appendix
%\section{Appendix}
\input{tex/appendix_LLM_implementation_details}
\input{tex/appendix_entityType_QATemplate}

\input{tex/appendix_dataset_details}
\input{tex/appendix_details_baselines}
\input{tex/appendix_frequency_wise_ablation_analysis}
\input{tex/app_mid_vs_High_PerformanceAnalysis}

\input{tex/app_why_semantic_variants_improves_retrieval}

\end{document}

%% file: tex/abstract.tex
Large language models (LLMs) struggle with relation completion (RC), both with and without retrieval-augmented generation (RAG), particularly when the required information is rare or sparsely represented. 
To address this, we propose a novel multi-stage paraphrase-guided \textit{relation-completion RAG} framework, \textbf{\textit{\sys{}}}, that systematically incorporates relation paraphrases across multiple stages. In particular, \textit{\sys{}}: (a) integrates paraphrases into retrieval to expand lexical coverage of the relation, (b) uses paraphrases to generate relation-aware summaries, and (c) leverages paraphrases during generation to guide reasoning for relation completion.
Importantly, our method does not require any model fine-tuning. Experiments with five LLMs on two benchmark datasets show that \textit{\sys{}} consistently outperforms several RAG baselines. In long-tail settings, the best-performing LLM augmented with \textit{\sys{}} improves by 40.6 Exact Match (EM) points over its standalone performance and surpasses two strong RAG baselines by 16.0 and 13.8 EM points, respectively, while maintaining low computational overhead.

%% file: tex/introduction.tex
Relation completion (RC) has emerged as an important task for large language models (LLMs), as it enables models to infer and reason about relations between entities. This capability supports a variety of applications, including knowledge graph completion and multi-hop reasoning, where LLMs are increasingly used as a central reasoning component \cite{wei-etal-2023-kicgpt,yao2025exploringlargelanguagemodels}. Although LLMs perform well on many tasks, they still struggle with relation completion. In particular, LLM performance deteriorates significantly when the required knowledge is rare or lies in the long tail of the data distribution. In such sparse settings, retrieval-augmented generation (RAG) \cite{guu2020realmretrievalaugmentedlanguagemodel, lewis2021retrievalaugmentedgenerationknowledgeintensivenlp, izacard2022atlasfewshotlearningretrieval} is used to provide external evidence to LLMs. While RAG has shown strong performance across many NLP tasks, experimental results indicate that SOTA RAG approaches still perform poorly for relation completion. This issue becomes even more severe when the underlying relations are rare or occur in long-tail settings. 

Formally, given a triple $(e_1, r, , ?)$, where $e_1$ denotes the head entity and $r$ denotes the relation, the goal of relation completion (RC) is to predict the missing tail entity $e_2$ to complete the triple $(e_1, r, e_2)$. %For example, given the query $(\textit{Milton Ottey}, \texttt{place of birth}, , ?)$, the RC system should produce the answer $(\textit{Milton Ottey}, \texttt{place of birth}, \textit{May Pen})$. In LLM-based approaches, the model predicts the missing entity by leveraging its internal knowledge or external evidence retrieved from a corpus through RAG.
We define a triple as long-tail\footnote{We acknowledge that several other scenarios may fall under the ``long-tail'' umbrella. We leave the development of a comprehensive taxonomy of long-tail relations for future work.} if the co-occurrence frequency of the entity pair $(e_1, e_2)$ within the same sentence of a passage in the corpus is below a threshold $x$. Under this definition, most triples $(e_1, r, e_2)$ have very limited evidence of the target relation in the corpus (see Figure~\ref{fig:frequency_histogram} in Appendix~\ref{app:dataset_details}). For example, the pair $(\textit{Bruno Zevi}, \textit{Sapienza University})$ % ms: let's use texttt for code and textit for text
appears only three times in the corpus used in this work. When conditioned on the relation \texttt{educated at}, the available evidence becomes even more sparse. Moreover, relations can be expressed through diverse paraphrases. For instance, \texttt{educated at} may be expressed in text using the paraphrase \texttt{alma mater}. % ms: let's use texttt for code and textit for text
If such variations are not considered during retrieval, relevant evidence may not be retrieved.

In this paper, we propose a novel \textit{relation-completion RAG} framework, \textbf{\textit{\sys{}}}. 
The novelty of our framework lies in incorporating relation paraphrases across multiple stages. First, they are infused into a hybrid retrieval mechanism to expand lexical coverage of relation expressions. This enhances the system's capacity to retrieve relevant evidence even when explicit mentions of the relation are sparse. The same paraphrases are further leveraged to produce relation-aware summaries that reduce noise and guide the model toward relevant evidence. Finally, they are used during generation to focus the model’s attention on relation-specific cues, enabling stronger reasoning for relation completion, without requiring any model fine-tuning.

%In relation completion, relevant evidence may be expressed using semantic variations that differ from the target relation, which can lead to retrieval failures. For example, when the relation $r$ is \texttt{founded by}, supporting evidence may instead use expressions such as \texttt{established by} or \texttt{created by}. To address this issue, we encode relation semantics through paraphrases of the target relation in the \textbf{semantic coverage retrieval} module, enabling the retriever to capture relevant evidence even when explicit mentions of the target relation are sparse. In the \textbf{evidence aggregation} stage, our framework leverages relation paraphrases to assign higher importance to sentences containing these paraphrases and to prioritize evidence mentioning entities consistent with the expected entity type, while filtering out distracting information to produce an evidence-aware summary. Finally, in the \textbf{relation-constrained generation} stage, the framework incorporates relation paraphrases to guide the model’s attention toward relationally relevant sentences in the summary, encouraging the model to focus on evidence that aligns with the semantic intent of the relation and improving its ability to infer the correct tail entity.

To our knowledge, this work provides the first systematic analysis of RAG performance on relation completion (RC) across frequency regimes, from high-frequency to long-tail relations. Our analysis reveals how sparsity affects retrieval quality and motivates the design of methods to address this challenge. %Our analysis reveals how different LLMs and existing RAG approaches struggle with RC and experience significant performance degradation when evidence is sparse. To address this challenge, we propose \sys{}, a novel RAG framework specifically designed for the RC task. Experiments show that \sys{} consistently outperforms existing RAG methods, achieving robust performance under sparse evidence and establishing a new state of the art across the entire frequency spectrum for RC.

All in all, our contributions are:

{\flushleft {\bf (1)}} We propose a novel RAG framework, \textbf{\emph{\sys{}}}, for relation completion that incorporates relation paraphrases across multiple stages of the framework. These paraphrases expand lexical coverage during retrieval, help identify relevant information during evidence summarization, and guide reasoning during generation, addressing the challenge of relation sparsity. To our knowledge, we are first to explore the impact of paraphrases for RC under RAG.
%{\flushleft {\bf (1)}} We propose \textbf{\emph{\sys{}}}, a RAG framework for relation completion that expands semantic coverage across three stages: \textbf{(a) semantic coverage retrieval}, \textbf{(b) evidence aggregation}, and \textbf{(c) relation-constrained generation}. By encoding semantic variations of the target relation across three stages, \sys{} expands semantic coverage and robustly retrieves and prioritizes relevant context even when relational mentions are rare, resulting in significant improvements in long-tail RC.

{\flushleft {\bf (2)}} We evaluate \textit{\sys{}} on two benchmark datasets (MALT and WikiData5M) across five LLMs, demonstrating consistent improvements over strong state-of-the-art RAG baselines. Importantly, in the long-tail setting, our framework surpasses two competitive RAG baselines by 16.0 and 13.8 EM points, respectively. \textit{\sys{}} requires no model fine-tuning, making it broadly applicable across different LLMs. %The best-performing LLM augmented with \textit{\sys{}} achieves a gain of 40.6 EM points over its standalone performance.

{\flushleft {\bf (3)}} This work is the first to conduct a systematic frequency-aware analysis of RAG performance for relation completion. Our results show that existing methods degrade sharply in long-tail settings, while our framework remains robust, providing an initial step toward addressing long-tail challenges in relation completion.

{\flushleft {\bf (4)}} We introduce a reproducible evaluation framework in which RAG baselines and \textit{\sys{}} operate under identical settings, using the same retrieval corpus, benchmark datasets, and evaluation metrics to ensure fair comparison\footnote{We will release all software and data upon acceptance.}.

%% file: tex/RAGImage.tex
\begin{figure*}[t]
  \centering
\includegraphics[width=0.95\textwidth]{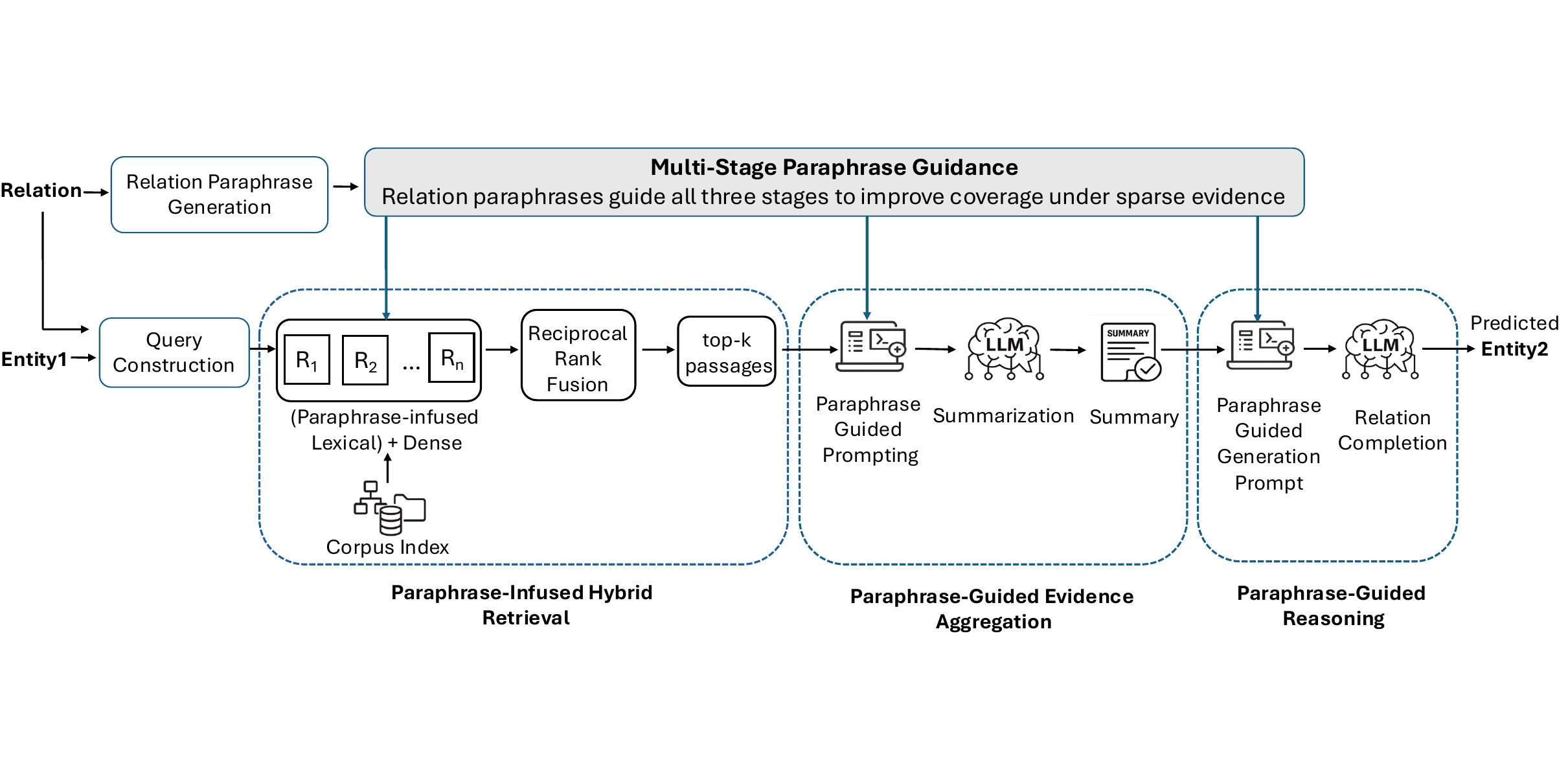}
\setlength{\abovecaptionskip}{-8pt}
 \caption{Overview of our multi-stage paraphrase-guided RAG framework, \textbf{\textit{\sys{}}} for relation completion. Given an entity–relation query, we generate relation paraphrases of the target relation and inject them across three stages: (1) paraphrase-infused hybrid retrieval to expand semantic coverage, (2) paraphrase-guided evidence aggregation to extract relation-relevant information, and (3) paraphrase-guided generation to constrain reasoning.}
  \label{fig:architecture}
\end{figure*}

%% file: tex/related_work.tex
\noindent \textbf{Frequency Bias and Memorization in LLMs: }
LLMs have been shown to store large amounts of factual knowledge in their parametric memory, enabling strong performance on a wide range of knowledge-intensive tasks without explicit retrieval mechanisms \cite{petroni-etal-2019-language, roberts-etal-2020-much}. However, LLMs perform poorly in relation completion, particularly when the required information is sparse. Recent studies show that parametric knowledge in LLMs is unevenly distributed and influenced by the frequency with which facts appear in the pretraining data. In particular, LLMs are more likely to recall facts frequently observed during pretraining, while rare or infrequent facts are harder to extract \cite{carlini2023quantifyingmemorizationneurallanguage, kandpal2023largelanguagemodelsstruggle}. This frequency-dependent memorization behavior poses a challenge in long-tail settings, where the required facts are rare or weakly represented in the data distribution. These limitations highlight the need for RAG frameworks that can retrieve relevant context even under sparse evidence conditions without requiring LLM fine-tuning.\\
\textbf{RAG for Relation Completion:}
RAG-based methods augment LLMs with external non-parametric memory by retrieving relevant documents at inference time and conditioning generation on the retrieved evidence \cite{guu2020realmretrievalaugmentedlanguagemodel, karpukhin2020densepassageretrievalopendomain, lewis2021retrievalaugmentedgenerationknowledgeintensivenlp,shuster2021retrievalaugmentationreduceshallucination,asai2023selfraglearningretrievegenerate, xu2023recompimprovingretrievalaugmentedlms, jiang-etal-2023-active, reichman2024densepassageretrievalretrieving}. This approach has been shown to improve factual accuracy and robustness across tasks such as open-domain question answering, fact verification, and knowledge-intensive reasoning \cite{khattab2020colbertefficienteffectivepassage, izacard-grave-2021-leveraging, ram2023incontextretrievalaugmentedlanguagemodels}. However, the application of RAG to RC remains relatively underexplored, especially in long-tail settings. In practice, the information required to complete a relation may be sparse or weakly expressed in the corpus, making it difficult for retrieval mechanisms to identify sufficient contextual evidence. These challenges highlight the need for RAG frameworks that remain robust across varying evidence frequencies and support relation completion under sparse evidence conditions.\\
\textbf{Architectural and Analytical Challenges of Current RAG Frameworks:}
Existing RAG frameworks are often architecturally complex and computationally expensive, relying on sophisticated machine learning components that require extensive fine-tuning, task-specific customization, and, in some cases, synthetic data generation for instruction tuning \cite{lin2024raditretrievalaugmenteddualinstruction, asai2023selfraglearningretrievegenerate, xu2023recompimprovingretrievalaugmentedlms}. Moreover, these methods operate as black-box systems, providing limited insight into model behavior. In particular, there is no systematic analysis of the sensitivity of RAG frameworks to data sparsity, making it difficult to diagnose failure modes or ensure robust performance across different frequency conditions. In this work, we take a first step toward addressing these challenges by studying retrieval performance under varying data frequency conditions.

%% file: tex/proposed_method.tex
We propose \textbf{\textit{\sys{}}}, a novel RAG framework for relation completion that consists of three stages: (1) paraphrase-infused hybrid retrieval, (2) paraphrase-guided evidence 
aggregation
, and (3) paraphrase-guided reasoning. Figure~\ref{fig:architecture} illustrates the framework design in detail. %The key idea of our approach is to capture semantic variations of the target relation and systematically incorporate them across all three stages of the framework.

\paragraph{Relation Paraphrases Generation:}
To capture lexical variations of relations, we construct a \textit{relation paraphrase set} for each target relation. For a relation $r$, we define a paraphrase set $P_r = \{p_1, p_2, \dots, p_k\}$, where each $p_i$ represents a natural language expression conveying the same relational meaning as $r$. For example, for the relation $r =$ \texttt{founded by}, the paraphrase set may include expressions such as \texttt{established by}, \texttt{created by}, and \texttt{started by}. This step is performed offline as a one-time preprocessing procedure (Figure~\ref{fig:architecture}). The complete list is provided in Table~\ref{tab:relational_phrases} in Appendix~\ref{app:LLMs_implementation_details} and is used across all three stages of \textit{\sys{}}. We generate this set for each relation using an LLM\footnote{We use {\tt gpt-4o-mini-2024-07-18}.} with the prompt template shown in Figure~\ref{fig:symKnowledge-genration-prompt} in Appendix~\ref{app:LLMs_implementation_details}. 

\subsection{Paraphrase-Infused Hybrid Retrieval}
\label{sec:sym_knowledge_generation}

In practice, relevant evidence in the RAG corpus may be expressed using linguistic variations that differ from the target relation, which can lead to retrieval failures. For example, when the relation $r$ is \texttt{educated at}, the corpus may contain information expressed using paraphrases such as \textit{alma mater}. This issue becomes more severe in long-tail scenarios, where the required evidence is scarce, and failing to capture such variations can significantly affect retrieval performance. To mitigate this issue, we  incorporate relation paraphrases into the hybrid retriever, enabling the retriever to capture relevant evidence even when explicit mentions of the target relation are sparse. To operationalize paraphrase-infused hybrid retrieval, we leverage both lexical and dense retrieval signals, incorporating relation paraphrases into the lexical retriever to expand lexical coverage during retrieval.

\subsubsection{Lexical Retrieval}

Lexical retrieval captures surface-level variations of a relation through explicit textual matches. To leverage this property, we augment the retrieval query with relation paraphrases that provide linguistic variants of the target relation. Given a head entity $e_1$ and relation $r$, we construct lexical queries using the paraphrase set $P_r=\{p_1,\dots,p_k\}$ and retrieve passages using a disjunction over conjunctions $(e_1 \wedge p_1) \vee \dots \vee (e_1 \wedge p_k)$. Incorporating paraphrases into the retrieval query increases the likelihood of retrieving relevant passages even when explicit mentions of the relation are scarce, and helps disentangle cases where the dense retriever retrieves passages describing the target relation for a different entity. We implement lexical retrieval using an inverted index built with Elasticsearch\footnote{\url{https://www.elastic.co/elasticsearch}}, ranking passages with the BM25 scoring function \cite{10.1561/1500000019}.

\subsubsection{Dense Retrieval}

Relevant evidence may also appear in implicit or semantically related formulations that do not share exact lexical matches with the target relation. To capture such cases, we complement lexical retrieval with dense retrieval, which models deeper semantic similarity between the query and corpus passages. Given an input $(e_1, r)$, we construct a query using a relation-specific question-answering (QA) template (see Table~\ref{tab:relation_qa_templates} in Appendix~\ref{app:qaTemplates_EntityTypes}) that encodes the semantics of the target relation. These templates are automatically generated using an LLM\footnote{We use {\tt gpt-4o-mini-2024-07-18}.}. For example, for the relation $r =$ \texttt{founded by}, the query becomes \textit{Who founded \{entity\}?}. The dense retriever operates over a Wikipedia passage corpus with precomputed Contriever embeddings\footnote{\url{https://dl.fbaipublicfiles.com/contriever/embeddings/contriever-msmarco/wikipedia_embeddings.tar}} generated using the Contriever model trained on MS MARCO \cite{izacard2022unsuperviseddenseinformationretrieval}, which is widely used in state-of-the-art RAG systems \cite{asai2023selfraglearningretrievegenerate, xu2023recompimprovingretrievalaugmentedlms}. We retrieve the top-$k$ passages using dense vector similarity.

%Finally, we combine the ranked results from both retrievers using Reciprocal Rank Fusion (RRF), which integrates complementary retrieval signals and produces a more robust ranking of candidate passages.

\subsubsection{Reciprocal Rank Fusion}
\label{sec:rrf}

We independently query the lexical and dense retrievers and combine their ranked outputs using Reciprocal Rank Fusion (RRF) \cite{10.1145/1571941.1572114}, an unsupervised rank aggregation method that prioritizes items highly ranked by any retriever and has been shown effective for combining heterogeneous retrieval signals in RAG settings \cite{islam-etal-2025-framework}. Given the set of retrievers $\mathcal{R}$ and the rank $\mathrm{rank}_r(p)$ of passage $p$ returned by retriever $r \in \mathcal{R}$, the RRF score is computed as:

\begin{equation}
\mathrm{RRF}(p) = \sum_{r \in \mathcal{R}} \frac{1}{c_{\mathrm{rrf}} + \mathrm{rank}_r(p)} .
\end{equation}

In all experiments, we set $c_{\mathrm{rrf}} = 0$ to emphasize top-ranked passages and select the top-$k$ passages based on the fused scores.

\subsection{Paraphrase-Guided Evidence 
Aggregation}
\label{sec:sym-guided-summarization}

After retrieval, the collected evidence often contains substantial noise and irrelevant information. When the retrieved context is long or loosely related to the target relation, LLMs may become distracted by irrelevant details, which can negatively affect downstream reasoning. To mitigate this issue, we guide the summarization stage using relation paraphrases and entity-type constraints. Specifically, sentences containing relation paraphrases are assigned higher importance, as they are more likely to express the target relation. In addition, we prioritize evidence that mentions entities consistent with the expected type of the target entity $e_2$. These signals help the model focus on relationally relevant information while filtering out distracting context.

We use an LLM\footnote{We use gpt-4o-mini-2024-07-18.} with the prompt shown in Figure~\ref{fig:prompt_summarization} to generate an evidence-aware summary of the retrieved passages. For each triple $(e_1, r, e_2)$, the expected type of the target entity $e_2$ is determined by the relation $r$. This mapping is manually constructed, as each relation corresponds to a small and well-defined set of plausible entity types (e.g., \texttt{GPE} or \texttt{LOC} for \texttt{place of birth}). The full mapping is provided in Table~\ref{tab:relation_expected_entity_types} in Appendix~\ref{app:qaTemplates_EntityTypes}.

\subsection{Paraphrase-Guided Reasoning}
\label{sec:generation}

During the generation stage, relation paraphrases guide the reasoning process. Even after the relation-aware summarization stage, the summary may still contain multiple entities and contextual details that are not directly relevant to the target relation. To address this challenge, we incorporate paraphrases of the target relation to guide the generator LLM’s attention toward relationally relevant sentences in the summary. This mechanism encourages the model to prioritize evidence aligned with the semantic intent of the relation, thereby improving its ability to identify the correct tail entity while reducing distractions from irrelevant context. We use five different LLMs (see Section~\ref{sec:LLMs}) as RC generators to predict the target tail entity $e_2$ given the input pair $(e_1, r)$, the relation-aware summary, and relation paraphrases that guide reasoning. The generation prompt is shown in Figure~\ref{fig:prompt_generation} in Appendix~\ref{app:LLMs_implementation_details}.

%% file: tex/experiments.tex
\subsection{Large Language Models for Generation}
\label{sec:LLMs}
We evaluate five open-source LLMs as relation completion (RC) generators: \llamaSevenB{}, \llamaThirteenB{}, \llamaEightB{}, \mistral{}, and \mixtral{}. To ensure a fair comparison, all models use the same generation prompt. No task-specific fine-tuning is performed; each model is used in its original pretrained or instruction-tuned form. Additional implementation details, prompts, and hyperparameters are provided in Appendix~\ref{app:LLMs_implementation_details}.

\subsection{RAG Settings}
\label{sec:rag-settings}

\subsubsection{RAG Corpus}
\label{sec:rag_corpus}
Following prior work \cite{asai2023selfraglearningretrievegenerate, karpukhin2020densepassageretrievalopendomain}, we use the same Wikipedia passage corpus\footnote{\url{https://dl.fbaipublicfiles.com/dpr/wikipedia_split/psgs_w100.tsv.gz}} for all retrieval components in our framework, as well as for all RAG baselines, to ensure a fair and consistent comparison.

\subsubsection{Top-$k$ Retrieval Settings}
We evaluate two retrieval settings based on the number of top passages $k$ used as context: Top-$10$ and Top-$20$. Baseline methods are reported under Top-$10$ only, while our framework is evaluated under both settings to assess the impact of increased contextual evidence using our design.

\subsection{Datasets}
\label{sec:datasets}
We evaluate our framework on two benchmark datasets: MALT \cite{chen-etal-2023-knowledge} and WikiData5M \cite{wang-etal-2021-kepler}. MALT is specifically designed for long-tail knowledge base completion, so we evaluate it using a balanced global subset without additional frequency partitioning. In contrast, WikiData5M spans a wide range of occurrence frequencies; therefore, we report both global results and frequency-based analyses on a curated subset to assess robustness across frequency regimes, from high-frequency to long-tail settings. Details on frequency-based partitioning, long-tail threshold selection, and dataset construction are provided in Appendix~\ref{app:dataset_details}.

%\paragraph{Long-tail Setting}
%We define a triple as long-tail if the co-occurrence frequency of the entity pair $(e_1, e_2)$ within the same sentence of a passage in the indexed RAG corpus is below a threshold $x$. Motivated by prior findings that facts appearing fewer than five times during training are harder for LLMs to recall \cite{carlini2023quantifyingmemorizationneurallanguage}, we analyze the distribution of entity-pair frequencies in WikiData5M within the RAG corpus . Figure~\ref{fig:frequency_histogram} in Appendix~\ref{app:dataset_details} shows that the dataset exhibits a highly skewed distribution toward low-frequency entity pairs, where many pairs occur fewer than five times and the number of instances steadily decreases as frequency increases. Based on this observation, we set the threshold to $x = 5$. The full analysis is provided in Appendix~\ref{app:dataset_details}.

\subsection{Baselines}
\label{sec:baselines}

\subsubsection{Parametric LLM Baselines (No Context)}
To assess how LLMs perform in relation completion across frequency bins, from high-frequency to long-tail, when relying solely on their parametric memory, we evaluate five LLMs (see Section~\ref{sec:LLMs}). We use the same generation prompt shown in Figure~\ref{fig:prompt_generation} in Appendix~\ref{app:baseline_rag_details}, but remove both retrieved contextual evidence and relation paraphrases. This setting isolates the models' parametric factual knowledge and enables analysis of performance degradation across frequency partitions, highlighting the impact of relational sparsity in long-tail settings even for strong pretrained LLMs.

\subsubsection{RAG Baselines}
We compare our RAG framework against two SOTA RAG baselines: SELF-RAG \cite{asai2023selfraglearningretrievegenerate} and RECOMP \cite{xu2023recompimprovingretrievalaugmentedlms}. Since both methods are primarily designed and evaluated for question answering (QA), we cast the relation completion task as a QA problem to enable a fair comparison. Both baselines rely on sophisticated and computationally intensive architectures that require substantial additional training beyond the base generator, including task-specific components and large-scale supervision. However, even when evaluated in the QA setting they are trained for, our experiments show that both baselines still degrade significantly in long-tail scenarios. Their architectural details are provided in Appendix~\ref{app:baseline_rag_details}.

\subsection{Evaluation Metrics}
\label{sec:evaluation_measures}
We evaluate relation completion using exact match (EM) and approximate match (AM) accuracy. EM requires an exact match to the gold tail entity; however, unlike MALT, WikiData5M provides only a single canonical surface form, penalizing semantically correct variants (e.g., \textit{Chevy Chase, Maryland} vs.\ \textit{Chevy Chase}). To mitigate surface-form mismatches, we adopt AM based on word-level Jaccard similarity with a $60\%$ threshold. Under this criterion, \textit{Chevy Chase, Maryland} is counted as correct ($2/3$ similarity\footnote{Punctuation is removed prior to Jaccard similarity computation.}), while \textit{Orlando, FL} is incorrect for \textit{Jacksonville, FL} ($1/3$ similarity).

%% file: tex/result_and_analysis.tex
\subsection{Overall Performance Across Datasets}
\label{sec:main_results}
We evaluate \sys{} with five LLMs under two context settings: Top-$10$ and Top-$20$. For baselines, we use \textsc{SELF-RAG}$_{7\mathrm{B}}$ (fine-tuned on LLaMA-2-7B) and RECOMP with \llamaSevenB{}, both evaluated under the Top-$10$ setting, with RECOMP using the same generation prompt as \textit{\sys{}}. The Top-$20$ setting is used only for \textit{\sys{}} to analyze the effect of deeper retrieved context.

\input{tables/malt_results_1k}
\input{tables/wikidata5M_results_1500}

\noindent Tables~\ref{tab:result_malt_full} and~\ref{tab:result_wikidata5M_full} summarize the performance of our proposed RAG framework on the MALT and WikiData5M datasets compared with multiple baselines. We draw the following key observations:

{\flushleft {\bf (1) Consistently Outperforms SOTA RAG Baselines.}} 
Our RAG consistently outperforms all baselines across all five LLMs under both Top-$10$ and Top-$20$ configurations on both datasets. On the MALT dataset, even with \llamaSevenB{} under Top-$10$, our RAG surpasses RECOMP and SELF-RAG by $21.5$ and $4.4$ EM points, respectively, while with \mistral{} under Top-$20$, our best-performing model, the margins increase to $25.6$ and $8.5$ EM points. A similar performance trend is observed on WikiData5M across models and context settings (shown in Table~\ref{tab:result_wikidata5M_full}).

{\flushleft {\bf (2) Strongly Outperforms LLM Baselines.}}
All LLMs perform poorly under the no-context baseline setting. Our RAG outperforms even the strongest no-context model (\mixtral{}), achieving absolute gains of $26.3$ and $27.5$ EM points under the Top-$10$ and Top-$20$ settings, respectively. This highlights the limitations of relying solely on parametric knowledge when relational evidence is sparse. A similar trend is observed on the WikiData5M dataset, where our RAG consistently surpasses all no-context baselines.

{\flushleft {\bf (3) Gains from Increased Retrieval Depth.}}
\textit{\sys{}} with the Top-$20$ variant consistently outperforms Top-$10$ on both datasets, indicating that additional retrieved evidence effectively mitigates relational sparsity in long-tail relation completion.

\subsection{Frequency-Based Performance Analysis on Wikidata5M}
%Results for 3 bins on wikidata5M 1500
\input{tables/wikidata5M_longtail-analysis}
Table~\ref{tab:result_wikidata5M_frequency} presents a frequency-based evaluation of our RAG framework, \sys{}, and all baselines on the Wikidata5M dataset.

{\flushleft {\bf (1)}} \textbf{Existing SOTA RAG methods degrade sharply as relation frequency decreases.} Despite their complex, custom-trained designs, both RECOMP and SELF-RAG exhibit substantial performance degradation from high-frequency to long-tail subsets. Under the Top-$10$ setting, RECOMP drops by $11.4$ EM points (from $52.2$ to $40.8$), while SELF-RAG degrades by $17.8$ EM points (from $60.8$ to $43.0$). In contrast, our RAG shows stronger robustness: even with the smallest LLM among the five, \llamaSevenB{}, the Top-$10$ setting yields a $3.0$ EM point improvement from high-frequency to mid-frequency and only an $8.0$ EM point degradation from high-frequency to long-tail. %\todo{This is a key observation. Keep the numbers here!}

{\flushleft {\bf (2)}} \textbf{Our best-performing Top-$10$ configuration consistently outperforms prior RAG methods across all frequency bins.} Under the Top-$10$ setting, our RAG with the strongest model (\mixtral{}) surpasses both RECOMP and SELF-RAG in high, mid, and long-tail subsets. In the high-frequency regime, it exceeds RECOMP and SELF-RAG by $15.2$ and $6.6$ EM points, respectively. The gains increase in the mid-frequency subset to $19.8$ and $16.6$ EM points, and remain largest in the long-tail setting, with improvements of $16.0$ over RECOMP and $13.8$ over SELF-RAG.

{\flushleft {\bf (3)}} \textbf{No-context baselines collapse under frequency sparsity, even with the strongest LLM.}  
All no-context baselines exhibit severe performance degradation as relation frequency decreases. Even the strongest no-context model, \mixtral{}, drops sharply from $57.2$ EM in the high-frequency subset to $38.4$ EM in the mid-frequency subset and further to only $16.2$ EM in the long-tail subset, corresponding to a total degradation of $41.0$ EM points. In contrast, when equipped with our RAG framework, \mixtral{} achieves $67.4$ EM (high-frequency), $69.2$ EM (mid-frequency), and $56.8$ EM (long-tail) under the Top-$10$ setting, effectively mitigating the catastrophic long-tail drop. %This comparison indicates that parametric knowledge alone is insufficient for long-tail relation completion and highlights that our neuro-symbolic RAG enables even the same LLM to generalize robustly across the full frequency spectrum.

{\flushleft {\bf (4)}} \textbf{Increasing retrieval depth consistently strengthens robustness under frequency sparsity.}
Expanding retrieval from Top-$10$ to Top-$20$ yields consistent gains across models, with the largest improvements in mid-frequency and long-tail subsets, demonstrating that additional contextual evidence is particularly beneficial when relational signals are sparse.

\paragraph{Analysis of Mid- vs High-Frequency Performance.}
Interestingly, the mid-frequency bucket slightly outperforms the high-frequency bucket in Table~\ref{tab:result_wikidata5M_frequency}. 
To investigate this pattern, we conduct two analyses. First, we perform a finer-grained frequency analysis by partitioning the dataset into 10 global frequency bins. The results show that performance within both the high- and mid-frequency regimes is not uniform: the first high-frequency bin achieves the highest EM overall, while performance declines in the extreme high-frequency bins due to increased ambiguity and heterogeneous contextual evidence. This indicates that the apparent mid $>$ high pattern is primarily driven by degradation in the highest-frequency bins rather than a consistent advantage of mid-frequency instances, indicating that the effect arises from data characteristics rather than systematic model behavior.

Second, we analyze the noise level of retrieved evidence in the high-frequency bin. A manual inspection of retrieved passages shows that high-frequency queries contain fewer passages explicitly expressing the target relation than mid-frequency queries, indicating higher retrieval noise. Detailed bin-level analysis, noise analysis, and qualitative examples are provided in Appendix~\ref{app:mid_vs_high_frequency}.

\subsection{Ablation Study}
We perform two ablation studies on WikiData5M: (i) a component-wise ablation and (ii) a frequency-wise ablation (see Appendix~\ref{app:freq_wise_ablation_study} for the latter).
\subsubsection{Component-Wise Ablation Study}
\input{tables/ablation_study_wikidata5m_full}
Table~\ref{tab:ablation_compact_full_wikidata5M} shows a component-wise ablation on Wikidata5M using \mistral{} (chosen over \mixtral{} for time efficiency) under Top-$10$\footnote{Top-$20$ shows similar trends.}. Each variant removes one component while keeping others fixed.

\paragraph{Ablation of Hybrid Retrieval.}
We analyze the contribution of hybrid retrieval by isolating the lexical and dense retrievers while keeping all other components fixed. \textit{Lexical-only} disables dense retrieval but retains paraphrase infusion across the framework, whereas \textit{dense-only} removes lexical retrieval and eliminates paraphrase-guided retrieval while preserving the use of paraphrases in later stages. Removing either retriever degrades performance compared to the full framework, confirming their complementarity. Dense-only performs worse than lexical-only, highlighting the importance of our paraphrase-guided retrieval.

\paragraph{Stage-wise Paraphrase-Guidance Ablations.}
We evaluate the role of paraphrase-guidance at each stage. To do so, we remove paraphrase-guidance from each of the three stages of \textit{\sys{}} individually and also perform a global ablation that removes paraphrase-guidance from all modules. Removing paraphrase-guidance at any stage reduces performance compared to the full framework, with the global ablation causing the largest degradation, underscoring the importance of our design.

{\flushleft {\bf Question 1:}} {\em How does paraphrase guidance improve retrieval?}\\
To better understand why paraphrases of the target relation improve retrieval performance, we analyze retrieval outputs by comparing lexical retrieval using only the relation name with retrieval augmented with relation paraphrases. Incorporating paraphrases into lexical queries enables the retrieval of substantially more passages that explicitly express the target relation, increasing relation-aligned retrieval from 19.5\% to 29.5\% (+10 percentage points). Detailed analysis is provided in Appendix~\ref{app:paraphrase_retrieval_analysis}.

{\flushleft {\bf Question 2:}} {\em Does increasing the number of relation paraphrases improve performance?}\\
We analyze the impact of paraphrase coverage by varying the number of relation paraphrases. On the Wikidata dataset with LLaMA-2-7B-chat, increasing the average number from $\sim$7 to $\sim$9 improves EM from 59.1 to 60.27. While the gain is modest, it is consistent, indicating that greater relation-specific paraphrase coverage contributes positively to performance and supports our design.

%% file: tables/malt_results_1k.tex
%MALT-1000-full-results-table
\begin{table}[h]
\centering
\small
\begin{threeparttable}
\renewcommand{\arraystretch}{0.95}
\setlength{\tabcolsep}{4pt}
\resizebox{\columnwidth}{!}{
\begin{tabular}{@{\extracolsep{2pt}}l l r r}
\toprule
\makecell{Context\\Length} & Model & \makecell{EM\\(\%)} & \makecell{AM\\(\%)} \\
\midrule

\multicolumn{4}{c}{\textbf{Baselines}} \\
\midrule

\multirow{4}{*}{No-context}
 & Llama-2-7B-chat-hf        & 13.6 & 17.0 \\
 & Llama-2-13B-chat-hf       & 16.0 & 19.4 \\
 & Meta-Llama-3.1-8B-Instruct        & 22.8 & 27.7 \\
 & Mistral-7B-Instruct-v0.3  & 16.1 & 20.1 \\
 & Mixtral-8x7B-Instruct-v0.1  & 34.7 & 38.4 \\

\\[-0.6ex]
\cdashline{1-4}
\\[-0.6ex]

\multirow{2}{*}{Top-10}
 & RECOMP                    & 37.1 & 39.0 \\
 & SELF-RAG                  & 54.2 & 57.3 \\

\midrule
\multicolumn{4}{c}{\textbf{Our RAG} (\sys{})} \\
\midrule

\multirow{4}{*}{Top-10}
 & Llama-2-7B-chat-hf        & 58.6 & 60.9 \\
 & Llama-2-13B-chat-hf       & 58.9 & 61.1 \\
 & Meta-Llama-3.1-8B-Instruct        & 59.3 & 61.3 \\
 & Mistral-7B-Instruct-v0.3  & 60.6 & 63.2 \\
 & Mixtral-8x7B-Instruct-v0.1  & 61.0 & 63.3 \\
\\[-0.6ex]
\cdashline{1-4}
\\[-0.6ex]

\multirow{4}{*}{Top-20}
 & Llama-2-7B-chat-hf        & 59.3 & 61.5 \\
 & Llama-2-13B-chat-hf       & 60.5 & 62.9 \\
 & Meta-Llama-3.1-8B-Instruct        & 60.3 & 62.6 \\
 & Mistral-7B-Instruct-v0.3  & \textbf{62.7} & \textbf{65.1} \\
 & Mixtral-8x7B-Instruct-v0.1  & 62.2 & 64.8 \\
\bottomrule
\end{tabular}
}
\end{threeparttable}
\caption{Performance of our RAG framework on the MALT dataset across five LLMs, including baselines.}

%\caption{Overall performance on the MALT dataset with 1{,}000 instances evaluated across five large language models. Results are reported using Exact Match (EM) and Approximate Match (AM). Bold values indicate the best performance among the compared methods.}

\label{tab:result_malt_full}
\end{table}

%% file: tables/wikidata5M_results_1500.tex
\begin{table}[h]
\centering
\begin{threeparttable}
\small
\renewcommand{\arraystretch}{0.95}
\setlength{\tabcolsep}{4pt}
\resizebox{\columnwidth}{!}{
\begin{tabular}{@{\extracolsep{2pt}}l l r r}
\toprule
\makecell{Context\\Length} & Model & \makecell{EM\\(\%)} & \makecell{AM\\(\%)} \\
\midrule

\multicolumn{4}{c}{\textbf{Baselines}} \\
\midrule

\multirow{4}{*}{No-context}
 & Llama-2-7B-chat-hf        & 21.0 & 23.4 \\
 & Llama-2-13B-chat-hf       & 21.9 & 23.7 \\
 & Meta-Llama-3.1-8B-Instruct        & 30.0 & 32.6 \\
 & Mistral-7B-Instruct-v0.3  & 25.7 & 27.5 \\
 & Mixtral-8x7B-Instruct-v0.1  & 37.3 & 40.7 \\
\\[-0.6ex]
\cdashline{1-4}
\\[-0.6ex]

\multirow{2}{*}{Top-10}
 & RECOMP                     & 47.5 & 50.9 \\
 & SELF-RAG                  & 52.1 & 55.3 \\

\midrule
\multicolumn{4}{c}{\textbf{Our RAG} (\sys{})} \\
\midrule

\multirow{4}{*}{Top-10}
 & Llama-2-7B-chat-hf        & 59.1 & 62.4 \\
 & Llama-2-13B-chat-hf       & 61.7 & 65.0 \\
 & Meta-Llama-3.1-8B-Instruct        & 61.1 & 64.6 \\
 & Mistral-7B-Instruct-v0.3  & 63.8 & 67.5 \\
 & Mixtral-8x7B-Instruct-v0.1  & 64.5 & 68.7 \\
\\[-0.6ex]
\cdashline{1-4}
\\[-0.6ex]

\multirow{4}{*}{Top-20}
 & Llama-2-7B-chat-hf        & 60.8 & 64.0 \\
 & Llama-2-13B-chat-hf       & 62.8 & 65.7 \\
 & Meta-Llama-3.1-8B-Instruct        & 62.0 & 65.3 \\
 & Mistral-7B-Instruct-v0.3  & \textbf{65.5} & \textbf{69.1} \\
 & Mixtral-8x7B-Instruct-v0.1  & \textbf{65.5} & 68.6 \\
\bottomrule
\end{tabular}
}
\end{threeparttable}
\caption{Performance of our RAG framework on the Wikidata5M dataset across five LLMs, including baselines}
%\caption{Overall performance on the Wikidata5M dataset with 1{,}500 instances evaluated across five large language models. Results are reported using Exact Match (EM) and Approximate Match (AM). Bold values indicate the best performance among the compared methods.}
\label{tab:result_wikidata5M_full}
\end{table}

%% file: tables/wikidata5M_longtail-analysis.tex
%\newcommand{\num}[1]{\makebox[2.4em][r]{#1}}
\begin{table*}[h]
\centering
%\small
\begin{threeparttable}
\scriptsize
\renewcommand{\arraystretch}{0.95}
\setlength{\tabcolsep}{1.0pt}
\resizebox{0.7\textwidth}{!}{
%\begin{tabular}{@{\extracolsep{2pt}}l l rr rr rr}
%\begin{tabular}{llcccccc}
\begin{tabular}{@{\extracolsep{2pt}}l l
S[table-format=2.2]@{\hspace{4pt}}S[table-format=2.2]
S[table-format=2.2]@{\hspace{4pt}}S[table-format=2.2]
S[table-format=2.2]@{\hspace{4pt}}S[table-format=2.2]}

\toprule
\makecell{Context\\Length} 
& Model
& \multicolumn{6}{c}{\textbf{Frequency}} \\
\cmidrule(lr){3-8}

& 
& \multicolumn{2}{c}{\makecell{High-frequency}} 
& \multicolumn{2}{c}{\makecell{Mid-frequency}} 
& \multicolumn{2}{c}{\makecell{Long-tail}} \\
\cmidrule(lr){3-4} \cmidrule(lr){5-6} \cmidrule(lr){7-8}

&
& \multicolumn{1}{c}{\makecell{EM\\(\%)}} & \multicolumn{1}{c}{\makecell{AM\\(\%)}}
& \multicolumn{1}{c}{\makecell{EM\\(\%)}} & \multicolumn{1}{c}{\makecell{AM\\(\%)}}
& \multicolumn{1}{c}{\makecell{EM\\(\%)}} & \multicolumn{1}{c}{\makecell{AM\\(\%)}} \\

%& 
%& \makecell{EM\\(\%)} & \makecell{AM\\(\%)} 
%& \makecell{EM\\(\%)} & \makecell{AM\\(\%)} 
%& \makecell{EM\\(\%)} & \makecell{AM\\(\%)} \\
\midrule

\multicolumn{8}{c}{\textbf{Baselines}} \\
\midrule

\multirow{4}{*}{No-context}
 & Llama-2-7B-chat-hf
 & \num{38.8} & \num{40.4} & \num{16.4} & \num{18.6} & \num{7.80} & \num{11.2} \\

 & Llama-2-13B-chat-hf
 & \num{41.8} & \num{43.6} & \num{17.4} & \num{18.6} & \num{6.60} & \num{8.80} \\
 & Meta-Llama-3.1-8B-Instruct        & \num{51.0} & \num{52.6} & \num{28.0} & \num{30.8} & \num{11.0} & \num{14.4} \\
 & Mistral-7B-Instruct-v0.3
 & \num{47.2} & \num{49.0} & \num{21.6} & \num{22.6} & \num{8.20} & \num{10.8} \\
 & Mixtral-8x7B-Instruct-v0.1  & \num{57.2} & \num{61.0} & \num{38.4} & \num{41.8} & \num{16.2} & \num{19.2} \\
\\[-0.6ex]
\cdashline{1-8}
\\[-0.6ex]

\multirow{2}{*}{Top-10}
 & RECOMP 
 & \num{52.2} & \num{56.6} & \num{49.4} & \num{51.4} & \num{40.8} & \num{44.8} \\

 & SELF-RAG 
 & \num{60.8} & \num{63.6} & \num{52.6} & \num{54.8} & \num{43.0} & \num{47.6} \\

\midrule
\multicolumn{8}{c}{\textbf{Our RAG} (\sys{})} \\
\midrule

\multirow{4}{*}{Top-10}
 & Llama-2-7B-chat-hf
 & \num{60.8} & \num{64.0} & \num{63.8} & \num{65.6} & \num{52.8} & \num{57.6} \\
 & Llama-2-13B-chat-hf
 & \num{63.8} & \num{67.0} & \num{66.8} & \num{69.0} & \num{54.4} & \num{59.0} \\
 & Meta-Llama-3.1-8B-Instruct        & \num{63.2} & \num{67.2} & \num{66.0} & \num{67.6} & \num{54.0} & \num{59.0} \\
 & Mistral-7B-Instruct-v0.3
 & \num{67.4} & \num{71.2} & \num{67.8} & \num{69.8} & \num{56.2} & \num{61.4} \\
 & Mixtral-8x7B-Instruct-v0.1  & \num{67.4} & \num{71.4} & \num{69.2} & \num{71.0} & \num{56.8} & \num{61.8} \\
\\[-0.6ex]
\cdashline{1-8}
\\[-0.6ex]

\multirow{4}{*}{Top-20}
 & Llama-2-7B-chat-hf
 & \num{62.2} & \num{65.2} & \num{65.2} & \num{67.4} & \num{55.0} & \num{59.4} \\
 & Llama-2-13B-chat-hf
 & \num{64.2} & \num{67.0} & \num{68.4} & \num{70.4} & \num{55.8} & \num{59.8} \\
 & Meta-Llama-3.1-8B-Instruct        & \num{62.4} & \num{65.6} & \num{68.2} & \num{70.0} & \num{55.4} & \num{60.2} \\
 & Mistral-7B-Instruct-v0.3
 & \textbf{67.8} & \textbf{71.4} & \num{70.2} & \num{72.4} & \textbf{58.6} & \textbf{63.4} \\
 & Mixtral-8x7B-Instruct-v0.1  & \textbf{67.8} & \num{71.0} & \textbf{70.8} & \textbf{72.6} & \num{57.8} & \num{62.2} \\
\bottomrule
\end{tabular}
}
\end{threeparttable}
\caption{Performance of our RAG across the full frequency spectrum on the WikiData5M dataset, with baselines.}
%\caption{Performance across the full frequency spectrum on the WikiData5M dataset, partitioned into high-frequency, mid-frequency, and long-tail subsets, each containing 500 instances. Results are reported using Exact Match (EM) and Approximate Match (AM), and bold values indicate the best-performing models. While baseline methods show progressively larger performance degradation from high-frequency to long-tail partitions, our RAG framework exhibits a substantially smaller drop.}

\label{tab:result_wikidata5M_frequency}
\end{table*}

%% file: tables/ablation_study_wikidata5m_full.tex
\begin{table}[h]
\centering
\begin{threeparttable}
\renewcommand{\arraystretch}{0.95}
\setlength{\tabcolsep}{4pt}
\resizebox{\columnwidth}{!}{
\begin{tabular}{@{\extracolsep{2pt}}l l r r}
\toprule
\makecell{Context\\Length} & Mistral-7B-Instruct-v0.3 & \makecell{EM\\(\%)} & \makecell{AM\\(\%)} \\
\midrule

\multicolumn{4}{c}{\textbf{Our RAG} (\sys{})} \\
\midrule

Top-10  & Full RAG framework                         & \textbf{64.8} & \textbf{68.7} \\
%Top-20  & Full system                         & 65.5 & 69.1 \\

\midrule

\multicolumn{4}{c}{\textbf{Ablation Study}} \\
\midrule

\multirow{5}{*}{Top-10}
 & Dense-only retrieval                     & 61.7 & 65.2 \\
 & Lexical-only retrieval                   & 62.6 & 66.0 \\
 & No relation paraphrases in any stage        & 61.2 & 64.9 \\
 & No relation paraphrases in summarization   & 62.7 & 66.0 \\
 & No relation paraphrases in retrieval       & 63.5 & 67.3 \\
 & No relation paraphrases in generation      & 64.0 & 67.5 \\

\bottomrule
\end{tabular}
}
\end{threeparttable}
\caption{Component-wise ablation results for Mistral-7B-Instruct-v0.3 on WikiData5M.}
%\caption{Compact ablation summary for Mistral-7B-Instruct-v0.3 on WikiData5M. Exact Match (EM) and Approximate Match (AM) are reported for the full system and ablated variants under the Top-$10$ retrieval setting, providing a condensed view of the frequency-wise results in Table~\ref{tab:ablation_wikidata5M_frequency_wise}.}

\label{tab:ablation_compact_full_wikidata5M}
\end{table}

%% file: tex/conclusion.tex
We present \textbf{\textit{\sys{}}}, a relation-completion RAG framework that incorporates relation paraphrases across multiple stages to improve performance under sparse relational evidence. By expanding lexical coverage during hybrid retrieval and leveraging paraphrases in relation-aware summarization and generation, our approach enables more effective and robust retrieval and reasoning without requiring model fine-tuning. Experiments across multiple LLMs and benchmark datasets show that \textit{\sys{}} consistently outperforms strong RAG baselines and significantly improves LLM performance over standalone models, with particularly large gains in long-tail settings.

%% file: tex/limitation.tex
Our study focuses on relation completion in English; it remains an open question whether the observed long-tail robustness generalizes to other languages.

%% file: tex/appendix_LLM_implementation_details.tex
\section{Large Language Models and Implementation Details}
\label{app:LLMs_implementation_details}

%We detail the LLMs used at each stage of our framework and describe their roles in symbolic knowledge generation, summarization, and relation completion.

\subsection{Implementation Details: Paraphrase Generation}

We use an LLM\footnote{We use {\tt gpt-4o-mini-2024-07-18}.} to generate relation paraphrases. The model is prompted in a zero-shot manner to act as a paraphrase generator for a given target relation. Figure~\ref{fig:symKnowledge-genration-prompt} illustrates the prompt used for paraphrase generation. Given a target relation, the prompt instructs the LLM to produce diverse linguistic expressions that convey the same relational meaning in natural text. Table~\ref{tab:relational_phrases} presents the full set of paraphrases for each relation across datasets.

 \begin{figure}[t]
  \centering
  \includegraphics[
    width=\columnwidth,
    trim=90 30 100 10,
    clip]{img/prompt-sym-know-gen.pdf}
  \vspace{-13mm}
    \caption{Example prompt used for relation paraphrase generation. The figure shows the input prompt and LLM-generated paraphrases for the \texttt{founded by} relation.}
  \label{fig:symKnowledge-genration-prompt}
\end{figure}

\input{tables/list_of_relational_phrases}

\paragraph{Practical Considerations.}
The relation paraphrase generation step requires less than one minute to generate and manually verify paraphrases for each relation. The LLM effectively produces relevant paraphrases for the target relations in our experiments. We do not impose a fixed number of paraphrases per relation, as different relations naturally exhibit varying degrees of linguistic diversity. In practice, this process produces between 4 and 11 relation-specific paraphrases, with an average of 7 per relation across the two datasets. The set can be iteratively expanded by adding new paraphrases. Our experiments show that increasing paraphrase coverage leads to improved performance.

\subsection{Implementation Details: Summarization Module}
\begin{figure}[t]
  \centering
  \includegraphics[
    width=\columnwidth,
    trim=150 50 170 10,
    clip
  ]{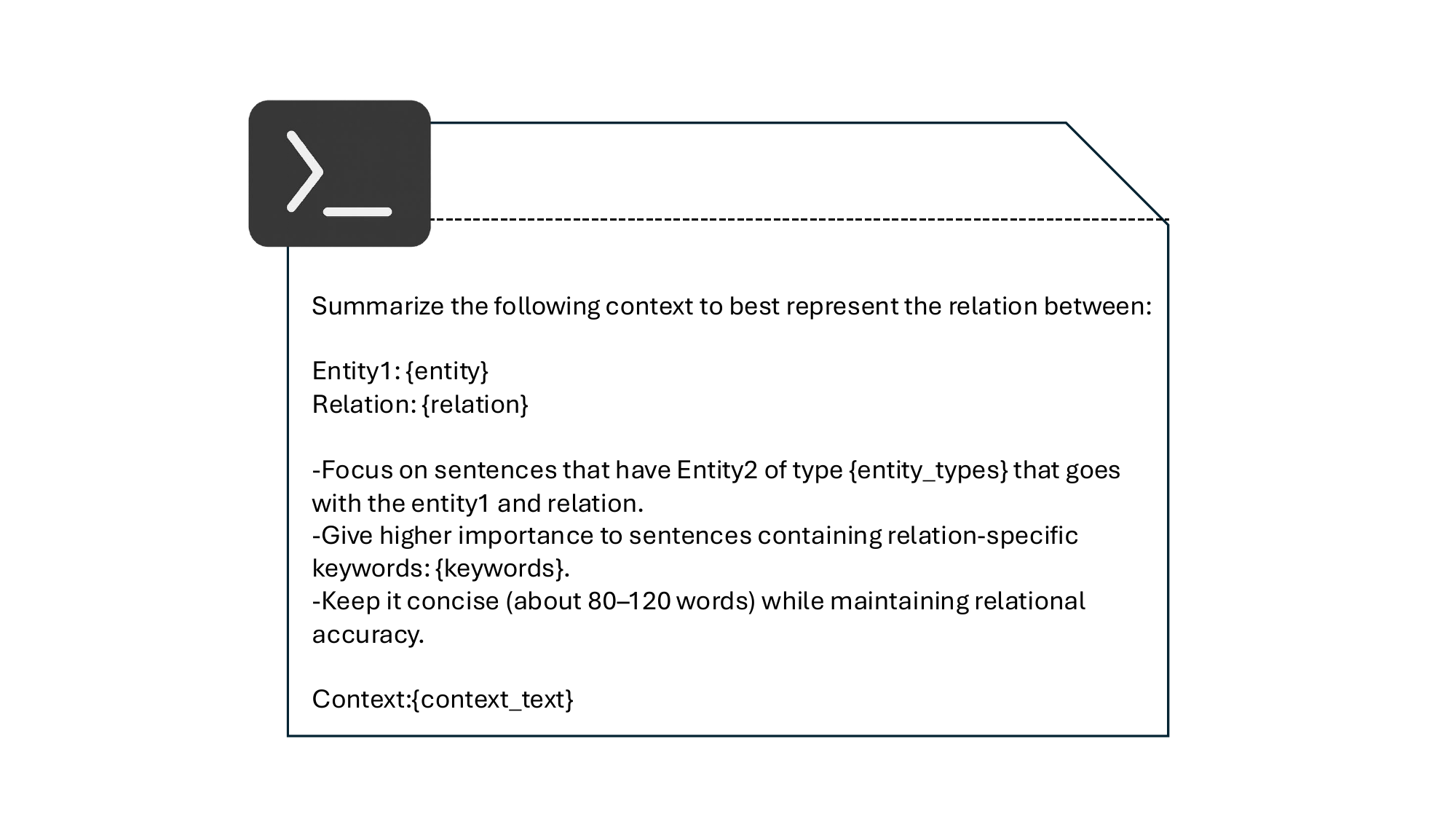}
    \caption{Prompt template used for relation-aware summarization with paraphrase guidance.}
  \label{fig:prompt_summarization}
\end{figure}

We use an LLM\footnote{We use {\tt gpt-4o-mini-2024-07-18}.} to generate relation-aware evidence summaries, following the methodology described in Section~\ref{sec:sym-guided-summarization}. For \textit{\sys{}}, we use the prompt template shown in Figure~\ref{fig:prompt_summarization}. No additional fine-tuning is performed for the summarization model.

\subsection{Implementation Details: Generation Module}
\begin{figure}[t]
  \centering
  \includegraphics[
    width=\columnwidth,
    trim=120 30 120 10,
    clip
  ]{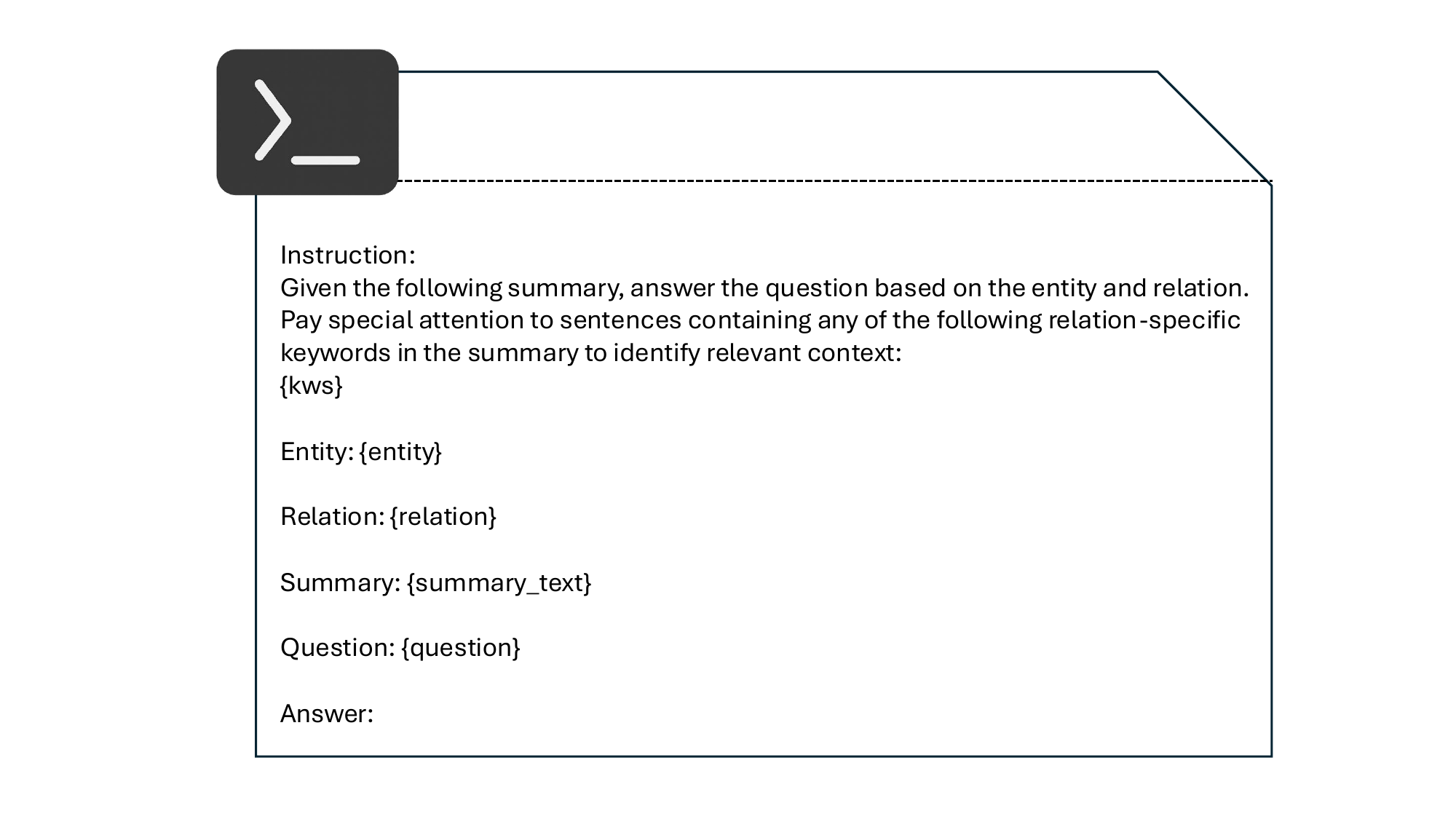}
\caption{Prompt template used for paraphrase-guided generation. We use the same prompt for all five LLMs in our experiments. Here, \texttt{kws} refers to the relation paraphrases incorporated during inference.}
  \label{fig:prompt_generation}
\end{figure}

In the generation module of our proposed RAG architecture (Figure~\ref{fig:architecture}), we perform relation completion using five different large language models: \llamaSevenB{}\footnote{\url{https://huggingface.co/meta-llama/Llama-2-7b-hf}}, \llamaThirteenB{}\footnote{\url{https://huggingface.co/meta-llama/Llama-2-13b-chat}}, \llamaEightB{}\footnote{\url{https://huggingface.co/meta-llama/Llama-3.1-8B-Instruct}}, \mistral{}\footnote{\url{https://huggingface.co/mistralai/Mistral-7B-Instruct-v0.3}}, and \mixtral{}\footnote{\url{https://huggingface.co/mistralai/Mixtral-8x7B-v0.1}}. We use the same generation prompt, shown in \autoref{fig:prompt_generation}, across all evaluated LLMs to ensure a fair and consistent comparison. No task-specific fine-tuning is performed; all models are used in their original pretrained form. We use chat- or instruction-tuned variants for all LLMs when available, as the generation prompt requires models to follow structured instructions, such as attending to relation paraphrases and conditioning on summarized contextual evidence.

\subsection{LLM Hyperparameters}

For relation-aware summarization, we use \gpt{} with a temperature of $0.0$; all other decoding parameters are left at their API defaults. For relation completion in the generation stage with open-source LLMs, we use greedy decoding (no sampling) with a temperature of $0.0$ and a maximum of 50 generated tokens. Input prompts are truncated to a maximum length of 2048 tokens.

%% file: tables/list_of_relational_phrases.tex
\begin{table*}[h]
\centering
\captionsetup{justification=centering}

\renewcommand{\arraystretch}{1.1}

\begin{adjustbox}{max width=\textwidth}
\setlength{\arrayrulewidth}{0.4pt}
\begin{tabular}{p{4.2cm}|p{10.8cm}}
\hline
\textbf{Relation} & \textbf{Relation Paraphrases} \\
\hline
founded by & founder, founded by, established by, created by, initiated by, was founded \\
\hline
performer & performed by, performer, singer, musician, artist, actor, actress \\
\hline
composer & composed by, composer, written by, music by, score by\\
\hline
place of birth & born, born in, birthplace, was born, place of birth\\
\hline
place of death & died, died in, passed away, place of death \\
\hline
employer & employed by, works for, worked for, job at, employer \\
\hline
educated at & studied at, educated at, graduated from, alma mater \\
\hline
residence & resides in, lives in, home in, residence, dwelling in \\
\hline
spouse$^{\bigstar}$ & spouse, married to, wife, husband, partner, wedded to, marriage to, life partner, spousal, their spouse \\
\hline
country of citizenship$^{\bigstar}$ & citizenship, citizen of, nationality, holds citizenship, country of origin, national of, belongs to, citizen status, legal nationality, national citizen, country of citizenship \\
\hline
shares border with$^{\bigstar}$ & borders, bordering, shares border, neighboring, adjacent to, common border, bordered by, touches, land border, nearby, shares border with \\
\hline
producer$^{\bigstar}$ & produced by, producer, film producer, music producer, executive producer, production by, producing, album producer, produced the, production credit \\
\hline
\end{tabular}
\end{adjustbox}

\caption{List of all relations in our MALT and Wikidata5M datasets and their associated relation paraphrases. Relations marked with $^{\bigstar}$ denote additional relations included only in the Wikidata5M subset.}
\label{tab:relational_phrases}
\end{table*}

%% file: tex/appendix_entityType_QATemplate.tex
\section{Relation-Specific QA Templates and Entity Type Constraints}
\label{app:qaTemplates_EntityTypes}

\subsection{Relation-Specific Templates}
Table~\ref{tab:relation_qa_templates} presents the list of relation-specific question–answering templates used for dense retrieval in our framework. The same templates are also used in \textit{\sys{}} during generation (see Figure~\ref{fig:prompt_generation} in Appendix~\ref{app:LLMs_implementation_details}) and for all state-of-the-art RAG baselines \cite{asai2023selfraglearningretrievegenerate, xu2023recompimprovingretrievalaugmentedlms} to ensure a fair comparison.

\input{tables/list_of_QA_template}
\subsection{Expected Tail Entity Type}
Table~\ref{tab:relation_expected_entity_types} shows the full list of expected tail entity types used during relation-aware summarization.
\input{tables/list_of_expected_entity_type}

%% file: tables/list_of_QA_template.tex
\begin{table}[h]
\centering
\captionsetup{justification=raggedright,singlelinecheck=false}

\renewcommand{\arraystretch}{1.1}

\begin{adjustbox}{max width=\columnwidth}
\setlength{\arrayrulewidth}{0.4pt}
\begin{tabular}{p{3.8cm}|p{4.4cm}}
\hline
\textbf{Relation} & \textbf{QA Template} \\
\hline
place of birth & Where was \{entity\} born? \\
\hline
place of death & Where did \{entity\} die? \\
\hline
founded by & Who founded \{entity\}? \\
\hline
composer & Who composed \{entity\}? \\
\hline
performer & Who performed \{entity\}? \\
\hline
employer & Who is \{entity\} employed by? \\
\hline
educated at & Where did \{entity\} study? \\
\hline
residence & Where does \{entity\} live? \\
\hline
spouse$^{\bigstar}$ & Who is the spouse of \{entity\}? \\
\hline
country of citizenship$^{\bigstar}$ & What country is \{entity\} a citizen of? \\
\hline
shares border with$^{\bigstar}$ & Which place does \{entity\} share a border with? \\
\hline
producer$^{\bigstar}$ & Who is the producer of \{entity\}? \\
\hline
\end{tabular}
\end{adjustbox}

\caption{Relation-specific question–answering templates. Relations marked with $^{\bigstar}$ are included only in the WikiData5M subset.}
\label{tab:relation_qa_templates}
\end{table}

%% file: tables/list_of_expected_entity_type.tex
\begin{table}[h]
\centering
\captionsetup{justification=raggedright,singlelinecheck=false}

\renewcommand{\arraystretch}{1.15}

\begin{adjustbox}{max width=\columnwidth}
\setlength{\arrayrulewidth}{0.35pt}
\begin{tabular}{p{3.8cm}|p{4.4cm}}
\hline
\textbf{Relation} & \textbf{Expected Tail Entity Type} \\
\hline
place of birth & GPE, LOC \\
\hline
place of death & GPE, LOC \\
\hline
founded by & PERSON \\
\hline
composer & PERSON \\
\hline
performer & PERSON, ORG \\
\hline
employer & ORG \\
\hline
educated at & ORG \\
\hline
residence & GPE, LOC \\
\hline
spouse$^{\bigstar}$ & PERSON \\
\hline
country of citizenship$^{\bigstar}$ & GPE, LOC \\
\hline
shares border with$^{\bigstar}$ & GPE, LOC \\
\hline
producer$^{\bigstar}$ & PERSON \\
\hline
\end{tabular}
\end{adjustbox}

\caption{Relation-specific expected tail entity types used to guide relation-aware summarization. Relations marked with $^{\bigstar}$ denote additional relations included only in the Wikidata5M subset.}
\label{tab:relation_expected_entity_types}
\end{table}

%% file: tex/appendix_dataset_details.tex
\section{Dataset Details}
\label{app:dataset_details}

We evaluate our framework on two datasets, MALT \cite{chen-etal-2023-knowledge} and WikiData5M \cite{wang-etal-2021-kepler}. For MALT, we perform evaluation on a balanced global subset without further partitioning, as the dataset is explicitly designed to emphasize long-tail relations. In contrast, for WikiData5M, we construct frequency-based partitions to analyze model performance under varying levels of relational sparsity.

\subsection{MALT}
%\todo{could be moved to an appendix}

The MALT dataset\footnote{\url{https://www.mpi-inf.mpg.de/de/departments/databases-and-information-systems/research/knowledge-base-recall/lm4kbc/malt}} (Multi-token, Ambiguous, Long-Tailed facts). was developed to benchmark knowledge base completion (KBC) methods specifically for long-tail relations, which often have limited facts in knowledge bases such as Wikidata. To construct MALT,  \citet{chen-etal-2023-knowledge} selected head entities from three broad categories, \texttt{business}, \texttt{music composition}, and \texttt{human}, which frequently involve long-tail entities.
 The dataset includes facts from eight different relations such as \texttt{founded by}, \texttt{performer}, capturing various relation structures. The full list of relations is given in the Table \ref{tab:relational_phrases}. The dataset statistically contains $87$\% of the sample facts with long-tail head entities, where a long-tail entity is defined as one having at most $13$ triples in Wikidata. Additionally, $65.3$\% of the dataset contains multi-word entities, and $58.6$\% of the facts involve ambiguous entities.
 Moreover, in the MALT dataset, each tail entity tends to have multiple valid paraphrases for the same answer. For example, for the tail entity \textit{Harvard University}, the gold annotations include multiple acceptable variations, such as \textit{Harvard University}, \textit{Harvard}, \textit{Harvard Graduate School}, and \textit{University of Harvard}, etc. Moreover, for most cases, relations such as \texttt{educated at} may have multiple correct tail entities, and the dataset explicitly accounts for this variability.
 
For the relation completion task, we construct a balanced subset of the original MALT dataset. To avoid bias toward or dominance by specific relations, we sample an equal number of instances per relation, with $125$ examples for each relation, resulting in a total of $1,000$ triples $(e_1, r, e_2)$.

\subsection{WikiData5M}
%\todo{could be moved to an appendix}

WikiData5M\footnote{\url{https://deepgraphlearning.github.io/project/wikidata5m}} is a knowledge graph (KG) benchmark dataset containing approximately 5 million entities, 20 million triples, 822 relations, and aligned entity descriptions from Wikipedia. The dataset provides both transductive and inductive evaluation settings. Since we do not perform any model training, we construct our subset from the transductive training split. We select the same eight relations included in the MALT dataset and additionally include four more relations to increase relational diversity, resulting in a total of twelve relations in our WikiData5M subset. The resulting subset contains 1,500 triples $(e_1, r, e_2)$.

\paragraph{Frequency-Based Partitioning.}
We partition instances based on the co-occurrence frequency of the entity pair $(e_1, e_2)$ within the same sentence of the same passage in the indexed RAG corpus. If an entity pair $(e_1, e_2)$ is already rare, conditioning on the relation $r$ further reduces its effective frequency in the corpus. Let $\mathit{count}(e_1, e_2)$ denote this frequency. We define three partitions: \textbf{long-tail} if $\mathit{count}(e_1, e_2) < x$, \textbf{mid-frequency} if $x \leq \mathit{count}(e_1, e_2) < 4x$, and \textbf{high-frequency} if $\mathit{count}(e_1, e_2) \geq 4x$, with each partition containing 500 examples.

Motivated by prior findings that facts appearing fewer than five times during training are difficult for LLMs to extract \cite{carlini2023quantifyingmemorizationneurallanguage}, we set $x = 5$. Within each partition, we apply approximately balanced sampling across the 12 relations by randomly allocating 41 or 42 examples per relation. In the high-frequency partition, some relations (e.g., \textit{residence}, \textit{composer}, \textit{educated at}, and \textit{producer}) contain insufficient examples; in such cases, remaining instances are randomly sampled from other relations to maintain 500 examples per partition. This partitioning enables a fine-grained analysis of how RAG methods behave as relational evidence becomes increasingly sparse.

\begin{figure}[t]
\centering
\includegraphics[
    width=\columnwidth,
    trim=0 4cm 0 0,
    clip
]{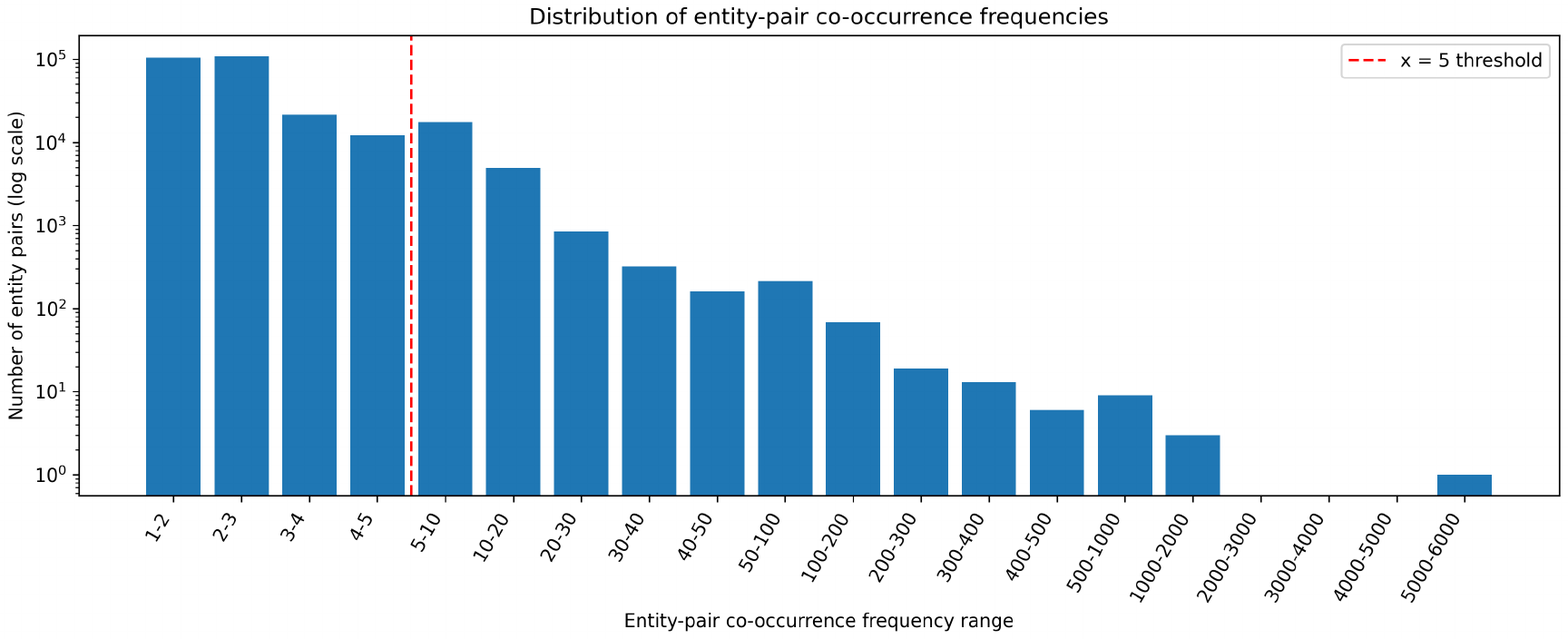}
\caption{Histogram of entity-pair co-occurrence frequencies in the indexed RAG corpus.}
\label{fig:frequency_histogram}
\end{figure}

\paragraph{Empirical Analysis for Long-Tail Threshold Selection}
While selecting the threshold value, we conducted an additional analysis of the entity-pair co-occurrence frequencies in the indexed corpus used for partitioning. The analysis shows (shown in Figure~\ref{fig:frequency_histogram}) that the distribution is highly skewed toward low-frequency entity pairs, with the many occurring fewer than five times and the number of instances steadily decreasing at higher frequencies. This confirms the long-tailed nature of the data and provides additional empirical motivation for our choice of the threshold. In particular, most examples lie in a data-sparse regime, and conditioning on a relation further reduces effective frequency. Therefore, using x=5 provides an intuitive boundary separating extremely sparse instances from more frequent ones. This choice is also consistent with prior findings showing that facts appearing fewer than 5 times during training are difficult for LLMs to extract.

%% file: tex/appendix_details_baselines.tex
\section{Details of Baseline RAG Methods}
\label{app:baseline_rag_details}
\subsection{SELF-RAG}
\label{sec:self_rag}
We include SELF-RAG \cite{asai2023selfraglearningretrievegenerate} as a strong state-of-the-art RAG baseline. SELF-RAG integrates retrieval, generation, and self-critique through extensive model training by extending a pretrained language model (LM) with reflection tokens, which are learned as part of standard next-token generation. The training framework follows a multi-stage process. First, a separate critic model based on LLaMA-2-7B is trained using supervision generated by prompting GPT-4 to annotate retrieval decisions and critique signals, which are then distilled into an in-house model. This critic is used offline to insert reflection tokens into training examples, enabling the generator to learn retrieval control, passage relevance assessment, factual grounding, and critique of intermediate generations. Subsequently, the final generator LMs (LLaMA-2-7B or LLaMA-2-13B) are trained on over 150K instruction-following instances augmented with retrieved passages and reflection tokens, using a conventional language modeling objective together with reinforcement learning-based optimization. This allows the generator to autonomously produce reflection tokens at inference time without relying on the critic. At inference time, SELF-RAG performs multi-stage retrieval, parallel passage processing, and critique-guided decoding with segment-level beam search. For retrieval, SELF-RAG uses the off-the-shelf Contriever-MS MARCO retriever \cite{izacard2022unsuperviseddenseinformationretrieval} and retrieves up to ten passages per query; we use the same retriever configuration in our experiments for a fair comparison. We run \textsc{SELF-RAG}$_{7\mathrm{B}}$, trained on LLaMA-2-7B, on both the MALT and WikiData5M datasets.

\subsection{RECOMP}
\label{sec:recomp}
We include RECOMP \cite{xu2023recompimprovingretrievalaugmentedlms} as another RAG baseline that improves generation by compressing retrieved passages before they are consumed by the generator. The method proposes two compression variants: an extractive compressor that selects salient spans from retrieved passages and an abstractive compressor that generates query-conditioned summaries. In our experiments, we use the abstractive summarization variant of RECOMP as a baseline, since our proposed framework also incorporates a summarization stage, enabling a fair comparison.

RECOMP first retrieves passages using a dense retriever; for fair comparison, we use the Contriever-MS MARCO retriever \cite{izacard2022unsuperviseddenseinformationretrieval}, consistent with SELF-RAG and our dense retrieval module. The abstractive compressor is implemented as an encoder---decoder sequence-to-sequence model initialized from T5 and pre-trained on a summarization dataset \cite{Hermann2015TeachingMT}. Its query-focused summarization ability is distilled from an extreme-scale language model by generating a large synthetic training dataset, filtering the generated summaries for quality, and training the encoder--decoder model on the filtered data. Training instances are constructed by pairing each retrieved passage with the input query and supervising the compressor to generate summaries that preserve answer-relevant evidence while discarding irrelevant content. The compressed summaries from multiple retrieved passages are concatenated and provided as context to the generator, reducing noise and effective context length. The compressor and generator are trained jointly using supervised learning.

%% file: tex/appendix_frequency_wise_ablation_analysis.tex
\section{Frequency-Wise Ablation Study}
\label{app:freq_wise_ablation_study}
\paragraph{Frequency-Wise Ablation Analysis on Wikidata5M.}
\input{tables/ablation_study_wikidata5M_frequency_wise}
Table~\ref{tab:ablation_wikidata5M_frequency_wise} analyzes the contribution of individual components across three frequency partitions. The full RAG framework consistently achieves strong performance across all regimes. Notably, in the long-tail setting, the lexical-only retrieval variant remains competitive even without dense retrieval, highlighting the effectiveness of relation paraphrase guidance in both retrieval and downstream stages. In contrast, removing paraphrase guidance from either retrieval (dense-only) or summarization results in clear performance degradation in the long-tail regime.

Interestingly, removing paraphrase guidance at the generation stage slightly improves performance in some cases by allowing the model to output multiple valid answers derived from the relation-aware summaries, particularly for relations that naturally support multiple correct entities (e.g., \textit{shares border with}, \textit{educated at}, \textit{producer}). In contrast, paraphrase guidance during generation encourages the model to produce a single, specific answer. Unlike MALT, WikiData5M contains incomplete gold annotations for such relations; consequently, correct but non-matching predictions are penalized. Removing paraphrase guidance during generation, which allows multiple answers to be produced, therefore increases the likelihood of matching the annotated gold labels, explaining the observed effect.

%% file: tables/ablation_study_wikidata5M_frequency_wise.tex
\begin{table*}[t]
\centering
\small
\begin{threeparttable}
\renewcommand{\arraystretch}{0.65}
\setlength{\tabcolsep}{4pt}
\resizebox{0.7\textwidth}{!}{
%\begin{tabular}{@{\extracolsep{2pt}}l l rr rr rr}
\begin{tabular}{@{\extracolsep{2pt}}l l
S[table-format=2.2]@{\hspace{4pt}}S[table-format=2.2]
S[table-format=2.2]@{\hspace{4pt}}S[table-format=2.2]
S[table-format=2.2]@{\hspace{4pt}}S[table-format=2.2]}

\toprule
\makecell{Context\\Length}
& (Mistral-7B-Instruct-v0.3)
& \multicolumn{6}{c}{\textbf{Frequency}} \\
\cmidrule(lr){3-8}

&
& \multicolumn{2}{c}{\makecell{High-frequency}}
& \multicolumn{2}{c}{\makecell{Mid-frequency}}
& \multicolumn{2}{c}{\makecell{Long-tail}} \\
\cmidrule(lr){3-4} \cmidrule(lr){5-6} \cmidrule(lr){7-8}

&
& \multicolumn{1}{c}{\makecell{EM\\(\%)}} & \multicolumn{1}{c}{\makecell{AM\\(\%)}}
& \multicolumn{1}{c}{\makecell{EM\\(\%)}} & \multicolumn{1}{c}{\makecell{AM\\(\%)}}
& \multicolumn{1}{c}{\makecell{EM\\(\%)}} & \multicolumn{1}{c}{\makecell{AM\\(\%)}} \\

%&
%& \makecell{EM\\(\%)} & \makecell{AM\\(\%)}
%& \makecell{EM\\(\%)} & \makecell{AM\\(\%)}
%& \makecell{EM\\(\%)} & \makecell{AM\\(\%)} \\
\midrule

\multicolumn{8}{c}{\textbf{Our RAG} (\sys{})} \\
\midrule
Top-10
 & Full RAG framework
 & \textbf{\num{67.4}} & \textbf{\num{71.2}} & \num{67.8} & \num{69.8} & \num{56.2} & \num{61.4} \\

\midrule
\multicolumn{8}{c}{\textbf{Ablation Study }} \\
\midrule

\multirow{7}{*}{Top-10}

 & Lexical-only retrieval
 & \num{66.0} & \num{68.6} & \num{65.2} & \num{68.4} & \textbf{\num{56.6}} & \num{61.0} \\
 &Dense-only retrieval
 & \num{65.8} & \num{69.6} & \num{65.2} & \num{68.2} & \num{54.2} & \num{57.8} \\
 & No relation paraphrases in retrieval
 & \num{66.2} & \num{70.2} & \num{67.8} & \num{70.0} & \num{56.6} & \textbf{\num{61.8}} \\

 & No relation paraphrases in summarization
 & \num{66.4} & \num{69.6} & \num{65.6} & \num{67.6} & \num{56.0} & \num{60.8} \\

 & No relation paraphrases in generation
 & \num{66.8} & \num{70.4} & \textbf{\num{69.0}} & \textbf{\num{71.0}} & \num{56.2} & \num{61.2} \\
 & No relation paraphrases in any stage 
 & \num{63.6} & \num{67.4} & \num{65.4} & \num{67.8} & \num{54.8} & \num{59.4} \\

\bottomrule
\end{tabular}
}
\end{threeparttable}
\caption{Ablation study analyzing the contribution of individual components of our RAG framework on the WikiData5M dataset, using Mistral-7B-Instruct-v0.3 as the underlying language model. Results are reported across high-frequency, mid-frequency, and long-tail partitions, highlighting the impact of our multi-stage paraphrase-guided RAG framework.}

\label{tab:ablation_wikidata5M_frequency_wise}
\end{table*}

%% file: tex/app_mid_vs_High_PerformanceAnalysis.tex
\section{Detailed Analysis of Mid- vs High-Frequency Performance.} 
\label{app:mid_vs_high_frequency}
\begin{figure}[t]
\centering
\includegraphics[
    width=\columnwidth,
    trim=0 9cm 0 0,
    clip
]{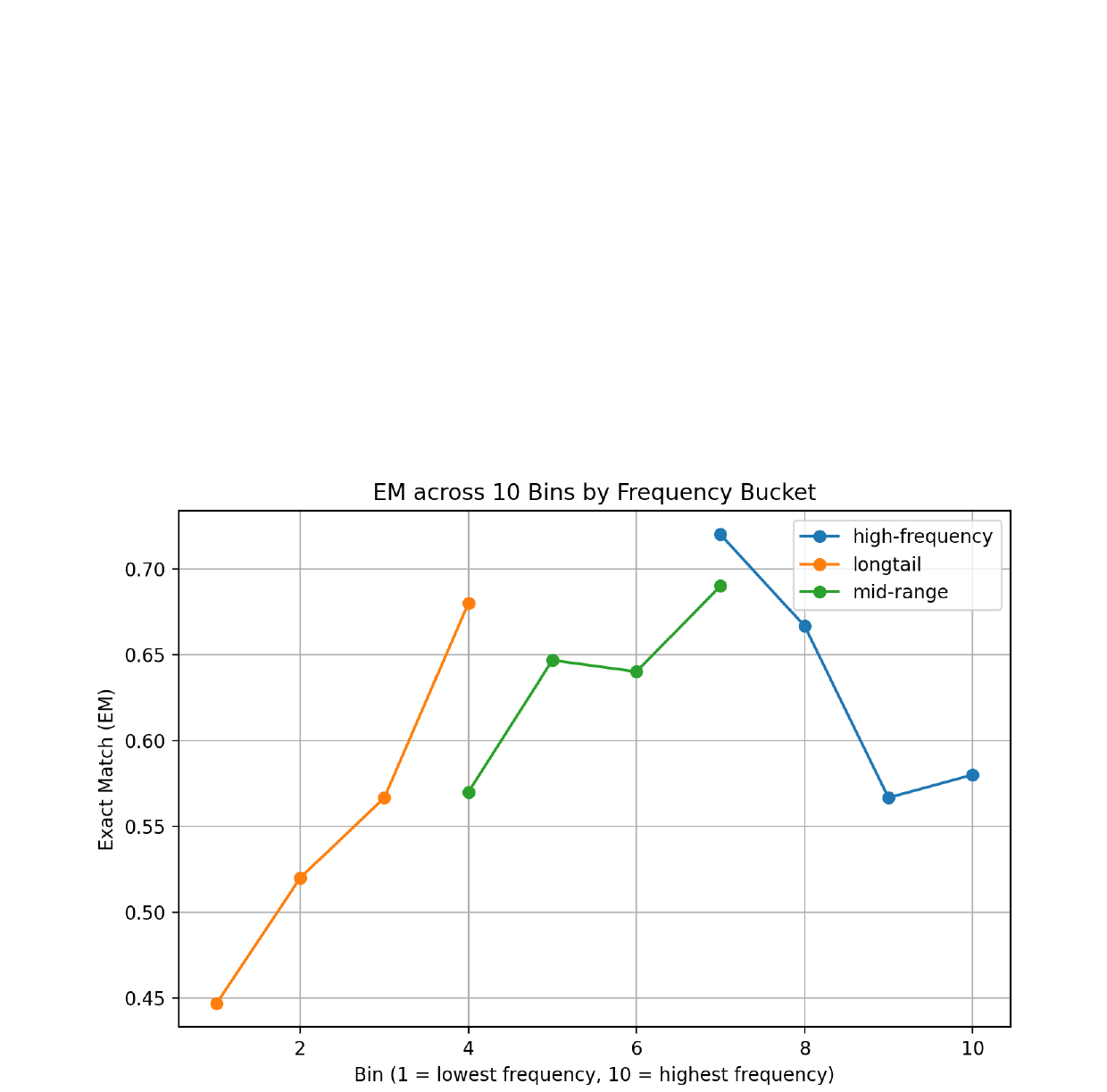}
\caption{Exact Match (EM) across 10 global frequency bins on WikiData5M.}
\label{fig:bucketwise_em_10bins}
\end{figure}

To better understand this phenomenon, we conduct a finer-grained analysis by dividing the Wikidata5M dataset into 10 equal-sized global bins sorted by entity-pair frequency. Examining the high-frequency bucket at this granularity reveals that bin 7 (shown in Figure~\ref{fig:bucketwise_em_10bins}) achieves the highest EM (0.72), outperforming all mid-frequency bins, while bin 8 (0.67) performs comparably to the strongest mid-frequency bin (0.69). However, performance drops substantially in bins 9 and 10. 

These findings indicate that the apparent mid $>$ high pattern is not due to a consistent advantage of mid-frequency instances or uniformly poor performance in the high-frequency bucket. Instead, performance varies across subranges, with degradation occurring primarily in the highest-frequency bins. This downward trend at extreme frequency levels is likely related to increased entity ambiguity and heterogeneous contextual evidence. Highly frequent entities tend to appear in more diverse contexts, often involving multiple plausible candidate entities, surface-form variations, and polysemous usages. As a result, although such entities have abundant evidence, the effective signal-to-noise ratio may decrease, leading to lower prediction accuracy.

\paragraph{Noise Level Analysis in the High-Frequency Bin.}
The observed performance drop in the highest-frequency bins further motivates an analysis comparing the noise level of retrieved passages in high- vs. mid-frequency buckets. To investigate this phenomenon, we manually analyze 20 instances from each bucket and count how many of the top-10 retrieved passages explicitly express the target relation. On average, mid-frequency instances contain 2.65 relevant passages per query, whereas high-frequency instances contain only 1.9. We define retrieval noise as the proportion of the top-10 retrieved passages that do not explicitly express the target relation (i.e., $1 - \text{relevant\_passages} / 10$). This corresponds to an estimated noise rate of 73.5\% for mid-frequency instances and 81\% for high-frequency instances.

A closer inspection reveals several factors contributing to the higher noise in the high-frequency bucket. First, entity ambiguity is more common among frequent entities. For example, in the high-frequency bin we observe ambiguous names such as \textit{Storni}, which may refer to either \textit{Alfonsina Storni} or \textit{Segundo Rosa Storni}. Retrieved passages often contain evidence for both entities, and the model tends to select the more dominant or frequently mentioned one. In such cases, the prediction is not necessarily factually incorrect but reflects name ambiguity. Similarly, instances such as \textit{Agis II} or \textit{Mustafa II} illustrate how retrieval may surface passages about closely related or more prominent entities (e.g., \textit{Mustafa I}), leading the model to favor a more popular or ambiguous variant that does not exactly match the gold entity.

Second, surface-form variations introduce additional mismatches (e.g., \textit{W E B Du Bois} vs.\ \textit{W.\ E.\ B.\ Du Bois} or \textit{chi lites} vs.\ \textit{Chi-Lites}), which may reduce exact lexical alignment despite semantic equivalence. Third, some entities exhibit strong polysemy across domains. For example, \textit{Tete} may refer to a song, a footballer, or a rodent species; highly frequent entities with multiple meanings tend to retrieve passages about the more popular interpretation rather than the gold target.

These observations suggest that high-frequency entities often retrieve abundant but heterogeneous information, increasing ambiguity and distractor signals. As a result, although such entities are frequent in the corpus, the effective signal-to-noise ratio for the target relation can be lower than in mid-frequency instances, helping explain the observed mid $>$ high pattern.

%% file: tex/app_why_semantic_variants_improves_retrieval.tex
\section{Why Infusing Relation Paraphrases Improves Retrieval}
\label{app:paraphrase_retrieval_analysis}

To better understand why lexical variants of the target relation improve retrieval performance, we analyze the retrieved passages under two lexical retrieval settings: (1) queries using only the canonical relation name and (2) queries augmented with relation paraphrases. We randomly sample 20 instances and examine the top-10 retrieved passages for each instance, resulting in 200 passages in total. For each passage, we manually determine whether it explicitly expresses the target relation.

Across the 20 sampled instances, relation-only lexical retrieval retrieves 39 passages (19.5\%; 1.95 per instance on average) that explicitly express the target relation. In contrast, lexical retrieval using relation paraphrases retrieves 59 such passages (29.5\%; 2.95 per instance on average), corresponding to a +10 percentage point increase in relation-aligned retrieval.

This improvement suggests that many relations are expressed in natural text through diverse linguistic forms that do not necessarily contain the canonical relation label. For example, relations such as \texttt{performer} or \texttt{employer} are often conveyed through verbs or prepositional phrases (e.g., \texttt{performed by}, \texttt{worked at}, \texttt{served as}), which may not be captured when querying only the relation name. By incorporating relation paraphrases, lexical retrieval better aligns with these natural language expressions, resulting in a higher proportion of retrieved passages that explicitly express the target relation.